\documentclass[sigconf]{acmart}
\AtBeginDocument{%
  }

\setcopyright{acmlicensed}
\copyrightyear{2018}
\acmYear{2018}
\acmDOI{XXXXXXX.XXXXXXX}
\acmConference[Conference acronym 'XX]{Make sure to enter the correct
  conference title from your rights confirmation email}{June 03--05,
  2018}{Woodstock, NY}
\acmISBN{978-1-4503-XXXX-X/2018/06}

\usepackage{multirow}

\begin{document}

\title{IMRouter: Imbalance-aware Temporal Pattern Routing for LLM-based Time Series Forecasting}

\author{Xin Yang}

\affiliation{%
  \institution{Xidian University}
  \city{Xi'an}
  \state{Shaanxi}
  \country{China}
}

\author{Cai Xu}

\affiliation{%
  \institution{Xidian University}
  \city{Xi'an}
  \state{Shaanxi}
  \country{China}
}

\author{Haishun Chen}

\affiliation{%
  \institution{Xidian University}
  \city{Xi'an}
  \state{Shaanxi}
  \country{China}
}

\author{Ying Liu}

\affiliation{%
  \institution{Xidian University}
  \city{Xi'an}
  \state{Shaanxi}
  \country{China}
}

\author{Jie Zhao}
\affiliation{%
  \institution{Chang'an University}
  \city{Xi'an}
  \state{Shaanxi}
  \country{China}
}

\author{Wei Zhao}
\affiliation{%
  \institution{Xidian University}
  \city{Xi'an}
  \state{Shaanxi}
  \country{China}
}

\author{Ziyu Guan}
\affiliation{%
  \institution{Xidian University}
  \city{Xi'an}
  \state{Shaanxi}
  \country{China}
}

\renewcommand{\shortauthors}{Trovato et al.}
\newcommand{\best}[1]{\textbf{#1}}
\newcommand{\second}[1]{\underline{#1}}

\begin{abstract}
Accurate time series forecasting (TSF) is essential for Web applications, such as traffic monitoring and online resource management. Recent advances in leveraging pre-trained large language models (LLMs) have shown promising performance in TSF. However, existing LLM-based TSF methods largely optimize the overall forecasting objective while overlooking temporal pattern imbalance (TPI), where different temporal patterns occur at highly unequal frequencies. Our empirical analysis reveals that TPI is prevalent across widely used TSF benchmarks, with substantially higher forecasting errors on less frequent patterns. To address this issue, we propose IMRouter (Imbalance-aware Temporal Pattern Routing), a novel framework that mitigates TPI-induced biases in both temporal representation learning and cross-modal alignment. Specifically, IMRouter introduces Pattern-Adaptive Temporal Representation Learning (PATR), which leverages gradient conflict and local gradient density to guide expert specialization. By penalizing incompatible expert assignments with density-aware weighting, PATR reduces the dominance of frequent optimization signals in router training. Furthermore, Routing-Guided Semantic Alignment (RGSA) transfers the pattern-specific knowledge learned by the router to cross-modal alignment through mutual information maximization, preserving distinctive temporal information across different patterns. Extensive experiments on benchmark datasets demonstrate the effectiveness of IMRouter, particularly in improving forecasting performance on less frequent temporal patterns.
\end{abstract}





\maketitle

\section{Introduction}
Time series forecasting (TSF) is essential for Web applications, including website traffic prediction, cloud workload management, and online service demand forecasting \cite{DBLP:conf/www/LiuHLDLHZ24,DBLP:conf/www/HuangZYW25}. Although statistical and deep learning methods \cite{taylor2018forecasting,DBLP:conf/aaai/ZhouZPZLXZ21} have achieved strong performance, their generalization across diverse temporal scenarios remains limited. With extensive knowledge acquired through large-scale pre-training, Large Language Models (LLMs), such as LLaMA \cite{grattafiori2024llama} and DeepSeek \cite{xu2026deepseek}, have shown promising generalization capabilities, motivating their application to TSF. However, the modality gap between discrete text and continuous time series remains a key challenge to effective forecasting \cite{DBLP:conf/kdd/JiangNPSNYSCNS25}.

\begin{figure}[t]
 \includegraphics[width=\linewidth]{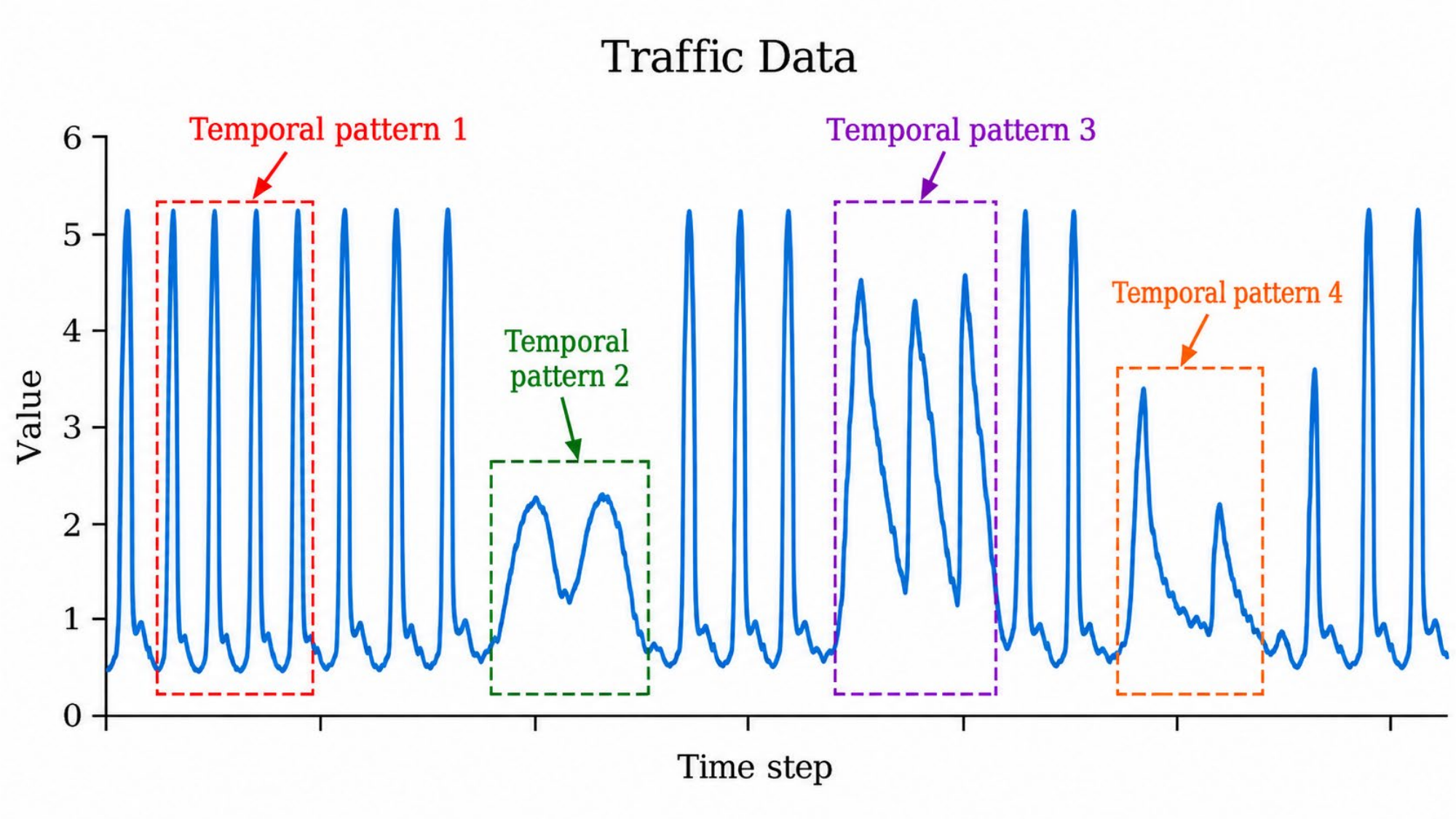}
  \caption{Illustration of temporal pattern diversity in a time series. Frequently recurring peak-shaped patterns coexist with less frequent, smoother local patterns (highlighted by red dashed boxes), demonstrating the heterogeneity of temporal dynamics.}
  \Description{Enjoying the baseball game from the third-base
  seats. Ichiro Suzuki preparing to bat.}
  \label{fig:pic1}
\end{figure}

To empower LLMs to explore time series and forecast accurately, most LLM-based TSF methods adopt the represent-then-align paradigm \cite{DBLP:conf/iclr/0005WMCZSCLLPW24}: (1) Representation: a temporal encoder is utilized to extract temporal representations from input time series; (2) Alignment: an alignment module takes the above temporal representations as queries to retrieve and aggregate text prototypes derived from the pre-trained vocabulary embeddings of LLMs; (3) Training: the temporal encoder and alignment module are optimized by minimizing the mean squared error (MSE) between the predicted and ground-truth future values, while the pre-trained LLM remains frozen.

\begin{figure*}
 \includegraphics[width=\textwidth]{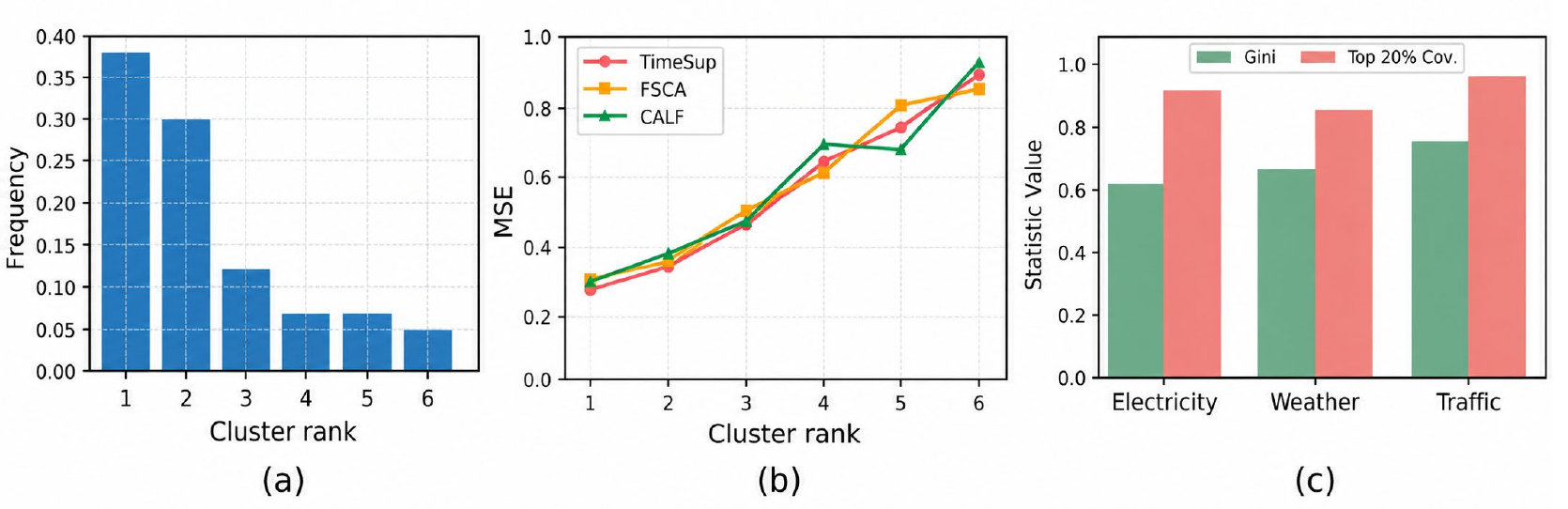}
  \caption{ (a) Frequency distribution of temporal pattern clusters identified from a representative channel in the Traffic dataset, ranked by cluster size in descending order. (b) Cluster-wise forecasting errors (MSE) of three existing methods on the same clusters, showing generally higher errors for less frequent patterns. (c) Dataset-level TPI statistics across Electricity, Weather, and Traffic. Gini measures the inequality of cluster sizes, while Top-20 \% Coverage denotes the proportion of samples covered by the largest 20 \% of clusters.}
  \Description{Enjoying the baseball game from the third-base
  seats. Ichiro Suzuki preparing to bat.}
  \label{fig:pic2}
\end{figure*}


Real-world time series exhibit diverse temporal dynamics, including trends, oscillations, and fluctuations \cite{DBLP:conf/iclr/WuHLZ0L23, DBLP:conf/iclr/WangWSHLMZ024}. We define a local temporal pattern as a recurring structure shared by segments with similar dynamics, as illustrated in Figure.~\ref{fig:pic1}. Inspired by class imbalance, we investigate whether such patterns also exhibit unequal frequencies. Since TSF datasets lack explicit pattern labels, we employ TS2Vec \cite{DBLP:conf/aaai/YueWDYHTX22} to encode fixed-length training windows and HDBSCAN \cite{10.1007/978-3-642-37456-2_14} to cluster similar representations. Each cluster approximates a local temporal pattern, with its size serving as a proxy for pattern frequency. As shown in Figure.~\ref{fig:pic2}(a), the cluster sizes are highly skewed, with a few dominant patterns and many less frequent ones. We refer to this phenomenon as Temporal Pattern Imbalance (TPI).

Under the MSE objective, frequently occurring temporal patterns contribute more samples and may dominate model optimization. This bias can affect LLM-based TSF methods in two ways. First, shared temporal encoders may favor frequent patterns, failing to capture the distinctive dynamics of less frequent ones. Second, cross-modal alignment may preferentially select semantic prototypes associated with frequent patterns, weakening the semantic representations of less frequent patterns. Consequently, forecasting performance on these patterns may deteriorate, as shown in Figure.~\ref{fig:pic2} (b). For example, Web services dominated by weekday workloads may fail to capture distinct weekend demand cycles, leading to inaccurate resource provisioning and degraded service quality. Addressing TPI is therefore important for reliable forecasting across diverse temporal patterns in Web applications.

A natural approach to modeling diverse temporal patterns is to employ a Mixture-of-Experts (MoE) architecture \cite{DBLP:journals/tnn/YukselWG12}, allowing different experts to specialize in distinct temporal dynamics. However, since the router is also optimized on the imbalanced training data, frequent patterns may dominate the routing optimization, biasing the router toward experts better suited to these dominant patterns. Consequently, less frequent temporal patterns may be assigned to experts primarily specialized for frequent patterns, preventing their distinctive temporal dynamics from being adequately captured. To this end, we propose Imbalance-aware Temporal Pattern Routing (IMRouter), a framework that mitigates TPI-induced biases in both temporal representation learning and cross-modal alignment. Specifically, IMRouter introduces Pattern-Adaptive Temporal Representation Learning (PATR), which leverages gradient conflict and local gradient density to guide expert specialization. By discouraging patches with conflicting gradients from sharing experts and emphasizing conflicts in low-density regions of the gradient space, PATR achieves more compatible expert assignments across diverse temporal patterns. Building on PATR, Routing-Guided Semantic Alignment (RGSA) transfers the learned routing information to cross-modal alignment. Specifically, it employs a mutual-information-based distillation objective to preserve pattern-discriminative information in the aligned representations, mitigating the dominance of frequent patterns in semantic prototype selection. The main contributions of this paper are summarized as follows:

\begin{itemize}
    \item We identify temporal pattern imbalance as a key challenge in LLM-based time-series forecasting, where frequent temporal samples dominate both representation learning and text prototype alignment.
    \item We propose IMRouter, a data-driven temporal-semantic balancing framework that mitigates head-dominated learning from both the representation and semantic alignment perspectives, without relying on manually defined head and tail labels.
    \item Extensive experiments on widely used TSF benchmarks demonstrate that IMRouter improves overall forecasting performance and enhances prediction accuracy on underrepresented temporal regimes.
\end{itemize}


\section{Related Work}
\subsection{LLM-based Time Series Forecasting}
Early works \cite{DBLP:conf/nips/ZhouNW0023}\cite{DBLP:conf/nips/GruverFQW23} simply represent time series as sequences of individual tokens and leverage LLMs for forecasting. However, they neglect the modality gap between numerical data and natural language. To mitigate this gap, Time-LLM \cite{DBLP:conf/iclr/0005WMCZSCLLPW24} introduces a reprogramming module that maps temporal representations to text prototypes derived from the pre-trained vocabulary. This transformation produces LLM-compatible representations, enabling the LLM to better model temporal dynamics. This paradigm has inspired subsequent studies on adapting LLMs to TSF tasks. CALF \cite{DBLP:conf/aaai/LiuG0LBR0X25} proposes a temporal–language consistency objective to strengthen cross-modal correspondence. FSCA \cite{DBLP:conf/iclr/0003LZYC25} explicitly models interactions between temporal and textual modalities through a graph neural network. TimeSup \cite{DBLP:conf/www/ZhengLZXMC26} reveals the pseudo-alignment problem from the perspective of data manifold dimensionality and mitigate this problem through dimension enhancement. Despite these advances, existing methods primarily focus on improving cross-modal correspondence while overlooking TPI. Consequently, frequent patterns may disproportionately influence both temporal representation learning and semantic prototype selection, potentially degrading forecasting performance on less frequent patterns.
\subsection{Imbalanced Time Series Forecasting}
Numerous efforts have been made to develop imbalance learning methods for classification tasks, such as cost-sensitive learning \cite{DBLP:journals/acisc/PetchpolB25, DBLP:conf/aaai/ChenXGZL25} and sampling techniques \cite{DBLP:conf/ijcai/ZhaoLQWRL022}. These methods fundamentally rely on class labels to identify imbalanced groups and apply different strategies to samples from different classes. Nevertheless, as mentioned above, TSF tasks lack explicit pattern labels, which makes it non-trivial to directly adapt these classification-oriented imbalance learning techniques to TSF tasks. Early imbalanced TSF approaches often rely on manually defined heuristics to assign imbalance-related labels to time-series samples, which allows the imbalance learning techniques developed for classification to be applied to TSF. For example, Silvestrin et al. \cite{DBLP:conf/mod/SilvestrinPH21} use the variation degree within a time window as an importance indicator to assign pseudo labels, and then apply resampling strategies to rebalance the sample distribution. Similarly, iBACon \cite{DBLP:journals/tkde/ZhangDY25} utilizes the discrepancy between input and output windows as a pseudo label. However, these heuristic indicators capture only specific aspects of temporal dynamics and cannot comprehensively characterize the imbalanced frequency distribution of diverse temporal patterns. Consequently, their effectiveness may depend on how well the predefined indicators reflect the underlying pattern distribution, limiting their generalizability across diverse forecasting scenarios.

TPI is related to, yet distinct from, several widely studied challenges in TSF, including abrupt changes, non-stationarity, and extreme events. While these challenges primarily concern sudden variations, evolving statistical properties, or unusually large deviations, TPI focuses on the unequal occurrence frequencies of recurring local temporal patterns. These phenomena may coexist but do not necessarily coincide. For example, less frequent patterns can exhibit smooth and recurring dynamics without abrupt changes or extreme values, and TPI may arise even in stationary time series. Consequently, methods developed for these related challenges do not necessarily mitigate the frequency-induced optimization bias addressed in this work.
\subsection{Mixture of Experts for Time Series Forecasting}
MoE-based forecasters employ specialized predictors to model different temporal dynamics. MoLE combines linear experts to adapt to changing temporal patterns \cite{DBLP:conf/aistats/Ni0WF24}. DUET routes series according to their latent temporal distributions \cite{DBLP:conf/kdd/QiuW0GH025}. Dynamic TMoE further adapts expert allocation to distribution drift \cite{zhu2026dynamic}. These methods rely on learned routers to identify relevant temporal characteristics and assign samples to suitable experts. However, the routers themselves are typically optimized on imbalanced training data, where frequent patterns may dominate routing decisions. Consequently, expert assignments may favor frequent patterns, potentially limiting effective specialization for less frequent ones.

Recent studies have explored clustering-guided expert specialization. TFPS employs subspace clustering to identify patch-level patterns for specialized experts \cite{DBLP:conf/nips/SunXECHW25}. By leveraging clustering-derived pattern information as additional guidance for expert assignments, TFPS can reduce reliance on the sample-averaged forecasting objective and partially mitigate the influence of TPI on router optimization. However, clustering results can be sensitive to algorithmic assumptions and hyperparameter choices, potentially affecting expert specialization. In contrast, IMRouter leverages gradient conflict and local gradient density to guide expert specialization directly from the model's optimization signals, reducing potential instability introduced by unreliable clustering results. IMRouter further transfers the learned pattern information to cross-modal alignment, providing a more comprehensive solution to TPI in LLM-based TSF.

\section{Preliminary}
Given a past time series $x^{time} \in \mathbb{R}^{C \times T_{past}}$, where $C$ denotes the number of variables and $T_{past}$ denotes the length of historical observations, the goal of LLM-based TSF methods is to predict the future values $\mathbf y^{\mathrm{time}} \in \mathbb{R}^{C \times T_{future}}$ over a prediction horizon $T_{future}$. The forecasting model can be formulated as:
\begin{equation}
\hat{y}^{time} = f_{\theta}(x^{time}),
\end{equation}
where $\hat{y}_{time}$ is the predicted future window and $\theta$ denotes the trainable parameters. 

Each channel $x^{time}_{c} \in \mathbb{R}^{1 \times T_{past}}$ of input time series $x^{time}$ is first normalized, divided into patches through a sliding window with size $p$ and stride $s$, and mapped into patch-level temporal tokens $e_c^n \in \mathbb{R}^{1 \times d}$:
\begin{equation}
\{e_c^n\}_{n=1}^{N}=\operatorname{PatchEmbed}(\operatorname{Normalize(x^{time}_{c})}),
\end{equation}
where $N = \lfloor \frac{T_{past}-p+s}{s} \rfloor$ denotes the number of patches.

To obtain LLM-compatible representations, the temporal embeddings are aligned with text prototypes through cross-attention:
\begin{equation}
z_{n}^{c}= \operatorname{MultiHeadAttention}(e_c^n,O,O),
\end{equation}
where $O \in \mathbb{R}^{P \times d_{model}}$ is the set of text prototypes obtained by compressing the pre-trained LLM vocabulary embedding matrix, and $P$ is the number of text prototypes. By treating temporal embeddings as queries and text prototypes as keys and values, this mechanism establishes interactions between numerical and linguistic representations.

The aligned representations are subsequently processed by the frozen LLM to generate forecasts $\hat{y}^{time}$. To optimize forecasting performance, the model is typically trained by minimizing the MSE loss:
\begin{equation}
\mathcal{L}(\theta)=\frac{1}{B}\sum_{b=1}^{B}||\hat{y}^{time}_b-y^{time}_b||^{2},
\end{equation}
where $B$ is the size of the mini-batch.

\begin{figure*}[t!]
 \includegraphics[width=\textwidth]{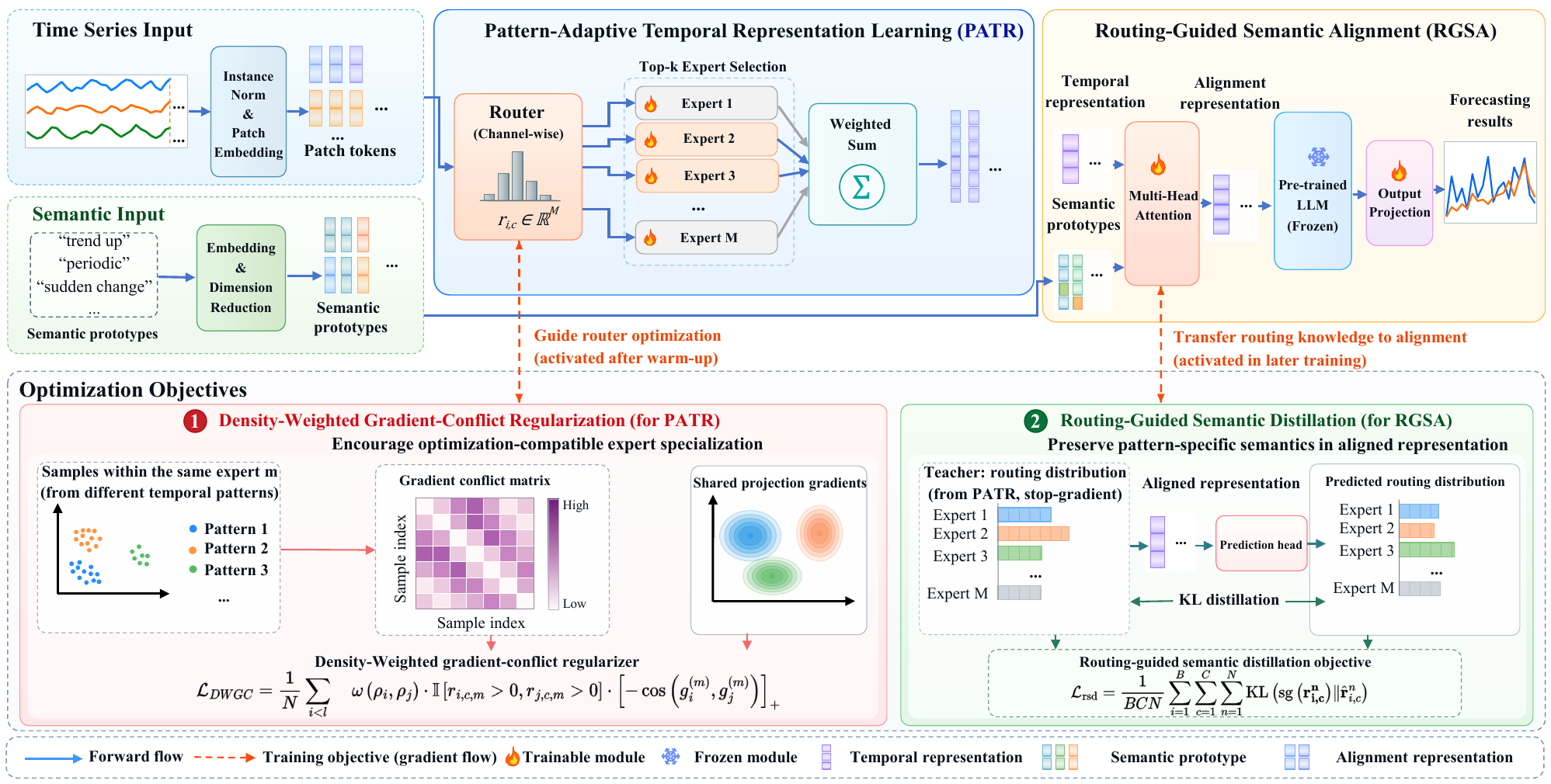}
  \caption{Illustration of IMRouter}
  \Description{Enjoying the baseball game from the third-base
  seats. Ichiro Suzuki preparing to bat.}
  \label{fig:pic3}
\end{figure*}

\section{ Method}
In this section, we begin by analyzing the MSE objective and show that the gradient contribution of each temporal pattern is implicitly weighted by its empirical frequency, leading frequent temporal patterns to dominate model optimization. Based on this observation, we propose IMRouter, a data-driven framework for imbalanced LLM-based TSF. As shown in Figure~\ref{fig:pic3} , IMRouter comprises two core components, PATR and RGSA, which jointly mitigate imbalance-induced bias in temporal representation learning and cross-modal semantic alignment.

\subsection{Motivation}
The standard training objective minimizes the average squared prediction error over all training samples, causing the optimization process to be dominated by the empirical data distribution. To support this argument, we decompose the MSE loss according to latent temporal patterns. Suppose the training dataset $D$ contains $J$ temporal pattern categories. The overall MSE objective can be rewritten as:
\begin{equation}
\mathcal{L}(\theta) = \sum_{j=1}^{J} \frac{|D_j|}{|D|}L_{j}(\theta),
\end{equation}
\begin{equation}
\mathcal{L}_j(\theta)=\frac{1}{|D_j|} \sum_{i \in \mathcal{D}_j} ||\hat{y}^{time}_i-y^{time}_i||^{2},
\end{equation}
where $|D|$ is the size of the training dataset, and $D_j$ denotes the set of samples belonging to temporal pattern category $j$ in the training dataset.

The gradient of the overall objective can be written as:
\begin{equation}
\nabla_\theta \mathcal{L}(\theta)=\sum_{j=1}^J \frac{|D_j|}{|D|} \nabla_\theta \mathcal{L}_j(\theta). 
\end{equation}

This decomposition shows that gradient contributions are weighted by empirical pattern frequencies, causing frequent temporal patterns to dominate optimization and rare samples to be insufficiently modeled.

\subsection{Pattern-Adaptive Temporal Representation Learning}

As discussed in Section 4.1, TPI may bias the optimization of shared temporal encoders toward frequent patterns. To mitigate this bias, PATR introduces a mixture-of-experts (MoE) architecture \cite{DBLP:journals/tnn/YukselWG12} that enables different temporal patterns to be processed by specialized experts. Since each expert is updated only by its assigned samples, this architecture can reduce optimization interference between patterns routed to different experts. However, the router of MoE is also trained on imbalanced data and may therefore favor frequent patterns. To address this limitation, PATR leverages gradient conflict to guide expert specialization, while local gradient density increases the relative importance of conflicts involving less densely represented optimization signals.

$\textbf{ Expert Routing}$ : 
Let $\mathbf e_{i,c}^{n}$ denote the patch embedding defined in Eq.~(2)
for the $n$-th patch of channel $c$ in sample $i$. To capture diverse temporal dynamics, PATR routes each patch to a sparse subset of $M$ experts. Specifically, the router produces an expert assignment distribution for each patch:
\begin{equation}
\mathbf r_{i,c}^{n}
=\operatorname{softmax}\!\left(g_{\phi}(\mathbf e_{i,c}^{n})\right),
\end{equation}
where $g_{\phi}$ is the routing network and
$\mathbf r_{i,c}^{n}\in\mathbb R^{M}$ represents the assignment probabilities over $M$ experts. The top-\(k\) experts are selected to construct the temporal representation :
\begin{equation}
\mathbf u_{i,c}^{n}
=\sum_{m\in\mathcal S_{i,c}^{n}}
\tilde r_{i,c,m}^{n}E_m(\mathbf e_{i,c}^{n}),
\end{equation}
where $\mathcal S_{i,c}^{n}=\operatorname{TopK}(\mathbf r_{i,c}^{n},k)$
is the selected expert set, $E_m$ is the $m$-th expert and $\tilde r_{i,c,m}^{n}$ denotes the normalized routing weight.

$\textbf{Density-aware Gradient Conflict Regularization}$ :
After an initial MoE warm-up, patches with similar temporal dynamics are expected to induce compatible expert-parameter gradients when assigned to the same expert. Opposing gradients may therefore indicate that the router has assigned patches with different optimization requirements to a shared expert. PATR leverages this signal to penalize their joint assignment probabilities to that expert, encouraging more compatible expert specialization.

Formally, let $q=(i,c,n)$ index a temporal patch, and
$\mathbf{g}_{q}^{(m)}$ denote its contribution to the
forecasting gradient of expert $m$. For two patches
$q$ and $q'$ assigned to the same expert, their
gradient conflict is measured as
\begin{equation}
\mathcal{C}_{q,q'}^{(m)}
=
\left[
-\operatorname{cos}
\left(
\mathbf{g}_{q}^{(m)},
\mathbf{g}_{q'}^{(m)}
\right)
\right]_{+},
\end{equation}
where $[x]_{+}=\max(0,x)$. The conflict score captures the degree of opposition between expert-parameter gradients, with positive values indicating potentially incompatible expert assignments.
Based on this measure, we formulate a gradient-conflict
regularizer that penalizes their joint assignment
probabilities:
\begin{equation}
\mathcal{L}_{\mathrm{GC}}
=
\frac{1}{|\mathcal{P}|}
\sum_{c,m}
\sum_{(q,q')\in\mathcal{P}_{c,m}}
\mathcal{C}_{q,q'}^{(m)}
r_{q,m}r_{q',m},
\end{equation}
where $\mathcal{P}_{c,m}$ denotes the set of sampled
patch pairs sharing expert $m$ within channel $c$,
$|\mathcal{P}|=\sum_{c,m}|\mathcal{P}_{c,m}|$,
and $r_{q,m}$ is the assignment probability of
patch $q$ to expert $m$.

Moreover, under severe TPI, conflicts involving less frequent patterns may be underrepresented in the batch-averaged objective, even when their individual conflict scores are high. To mitigate this dilution, we introduce inverse gradient-density weighting to enhance the relative influence of sparsely represented optimization signals. Since expert-specific gradient densities are influenced by routing assignments, they may not reflect the overall distribution of temporal patterns. We therefore estimate gradient density from the gradients of the shared router layer, enabling stable comparisons across temporal patterns.

Let $\mathbf{G}_{q}$ denote the contribution
of patch $q$ to the gradient of the shared embedding
parameters, and $\hat{\mathbf{G}}_{q}$ its normalized
form. Our weighting requires a direction-specific measure of update redundancy: gradients with similar directions indicate similar update demands. We therefore estimate the concentration around each normalized gradient using a Gaussian kernel
kernel:
\begin{equation}
\rho_q =
\frac{1}{|\mathcal{B}_c|}
\sum_{q'\in\mathcal{B}_c}
\exp\left(
-\frac{
\|\hat{\mathbf{G}}_q-\hat{\mathbf{G}}_{q'}\|_F^2
}{2\tau^2}
\right),
\end{equation}
where $\mathcal{B}_c$ denotes the set of patch
representations from channel $c$ within a mini-batch,
and $\tau$ is the kernel bandwidth.
The corresponding density-aware weight is defined as
\begin{equation}
w_q =
\frac{
(\rho_q+\epsilon)^{-\alpha}
}{
\frac{1}{|\mathcal{B}_c|}
\sum_{q'\in\mathcal{B}_c}
(\rho_{q'}+\epsilon)^{-\alpha}
},
\end{equation}
where $\alpha$ controls the reweighting strength.

Combining gradient conflict and local density,
we formulate the Density-Weighted Gradient Conflict
(DWGC) regularizer:
\begin{equation}
\mathcal{L}_{\mathrm{DWGC}}
=
\frac{
\displaystyle
\sum_{c=1}^{C}\sum_{m=1}^{M}
\sum_{(q,q')\in\mathcal{P}_{c,m}}
\omega_{q,q'}\mathcal{C}_{q,q'}^{(m)}
r_{q,m}r_{q',m}
}{
\displaystyle
\sum_{c=1}^{C}\sum_{m=1}^{M}
\sum_{(q,q')\in\mathcal{P}_{c,m}}
\omega_{q,q'}+\epsilon
},
\end{equation}
where $\mathcal{P}_{c,m}$ denotes the sampled pairs
from channel $c$ that share expert $m$,
$r_{q,m}$ is the assignment probability of patch
$q$ to expert $m$, and
$\omega_{q,q'}=\sqrt{w_qw_{q'}}$ represents their
joint density-aware weight.

\subsection{Routing-Guided Semantic Alignment}
During training, the PATR router learns to select the most suitable expert combination for each sample, thereby encoding rich temporal pattern information. Building on PATR, IMRouter further introduces RGSA to transfer pattern-aware knowledge from the PATR router to the aligned representation, preserving discriminative semantics for rare temporal patterns. Given the temporal representations $\{u_{i,c}^n\}_{n=1}^{N}$ generated by PATR, RGSA aligns them with the text prototypes through cross-attention:
\begin{equation}
a_{i,c}^n = \operatorname{MultiHeadAttention}(u_{i,c}^n, O, O).
\end{equation}

\begin{table*}[t]
\centering

\label{tab:long_term_forecasting}

\setlength{\tabcolsep}{3pt}
\renewcommand{\arraystretch}{1.1}

\resizebox{\textwidth}{!}{%
\begin{tabular}{l|cc|cc|cc|cc|cc|cc|cc|cc|cc}
\toprule
\multirow{2}{*}{Models}
& \multicolumn{2}{c|}{IMRouter}
& \multicolumn{2}{c|}{TimeSup}
& \multicolumn{2}{c|}{FSCA}
& \multicolumn{2}{c|}{CALF}
& \multicolumn{2}{c|}{Time-LLM}
& \multicolumn{2}{c|}{GPT4TS}
& \multicolumn{2}{c|}{PatchTST}
& \multicolumn{2}{c|}{iTransformer}
& \multicolumn{2}{c}{TimesNet} \\

& MSE & MAE
& MSE & MAE
& MSE & MAE
& MSE & MAE
& MSE & MAE
& MSE & MAE
& MSE & MAE
& MSE & MAE
& MSE & MAE \\
\midrule

ETTm1
& \underline{0.389} & \textbf{0.389}
& 0.402 & 0.407
& 0.473 & 0.437
& 0.393 & \underline{0.391}
& 0.410 & 0.409
& \underline{0.389} & 0.397
& \textbf{0.388} & 0.402
& 0.407 & 0.411
& 0.410 & 0.418 \\

\midrule

ETTm2
& \underline{0.279} & \textbf{0.321}
& 0.284 & 0.329
& 0.282 & 0.326
& 0.282 & \underline{0.322}
& 0.285 & 0.331
& \textbf{0.258} & 0.331
& 0.290 & 0.333
& 0.291 & 0.334
& 0.295 & 0.332 \\

\midrule

ETTh1
& 0.442 & \underline{0.437}
& 0.464 & 0.450
& \underline{0.429} & \textbf{0.436}
& \textbf{0.411} & 0.439
& 0.460 & 0.449
& 0.447 & \textbf{0.436}
& 0.448 & 0.447
& 0.464 & 0.455
& 0.460 & 0.455 \\

\midrule

ETTh2
& \textbf{0.372} & \textbf{0.397}
& 0.402 & 0.419
& \textbf{0.372} & \underline{0.401}
& \underline{0.376} & 0.404
& 0.389 & 0.408
& 0.381 & 0.408
& 0.384 & 0.413
& 0.383 & 0.407
& 0.407 & 0.421 \\

\midrule

Weather
& \underline{0.250} & \textbf{0.269}
& 0.259 & 0.279
& 0.260 & 0.280
& \textbf{0.246} & \underline{0.271}
& 0.274 & 0.290
& 0.264 & 0.284
& 0.258 & 0.280
& 0.260 & 0.281
& 0.259 & 0.285 \\

\midrule

ECL
& \textbf{0.171} & \textbf{0.263}
& 0.192 & 0.281
& 0.194 & 0.278
& 0.193 & 0.278
& 0.223 & 0.309
& 0.205 & 0.290
& 0.204 & 0.294
& \underline{0.175} & \underline{0.266}
& 0.197 & 0.297 \\

\midrule

Traffic
& \textbf{0.423} & \textbf{0.285}
& 0.442 & \underline{0.288}
& 0.457 & 0.293
& 0.459 & 0.303
& 0.541 & 0.358
& 0.488 & 0.317
& 0.482 & 0.308
& \underline{0.428} & \textbf{0.285}
& 0.625 & 0.331 \\

\bottomrule
\end{tabular}%
}
\caption{Long-term forecasting results averaged over four
prediction horizons on seven benchmarks. The best and
second-best results are shown in bold and underlined,
respectively.}
\end{table*}

To prevent rare-pattern information from being diluted during cross-modal alignment, RGSA encourages aligned semantic representations to preserve the routing information learned by PATR. Let $\mathbf{r}_{i,c}\in\mathbb{R}^{M}$ denote the routing distribution of the $i$-th sample and $c$-th channel, and $\mathbf{a}_{i,c}^{n}$ denote its aligned semantic representation. We aim to maximize the mutual information between the corresponding random variables $A$ and $R$:
\begin{equation}
I(A;R)=H(R)-H(R|A),
\end{equation}
where $R$ represents the expert-assignment variable induced by the routing distribution, and $A$ denotes the aligned semantic representation.

Since the true posterior $p(R|A)$ is intractable, we introduce a lightweight predictor $q_\psi$ to approximate it from the aligned representation:
\begin{equation}
\hat{\mathbf{r}}_{i,c}^{n}=\operatorname{softmax}\left(q_\psi\left(\mathbf{a}_{i,c}^{n}\right)\right).
\end{equation}
By the non-negativity of KL divergence, the mutual information admits the following variational lower bound:
\begin{equation}
\begin{aligned}
I(A;R)
&=H(R)+\mathbb{E}_{p(A,R)}[\log p(R|A)]\\
&\geq H(R)+\mathbb{E}_{p(A,R)}[\log q_\psi(R|A)].
\end{aligned}
\end{equation}
Using the routing distribution as a soft target, the negative expected log-likelihood can be estimated by the cross-entropy between $\mathbf{r}_{i,c}$ and $\hat{\mathbf{r}}_{i,c}^{n}$. Since cross-entropy decomposes into entropy and KL divergence, and the routing distribution is detached during distillation, maximizing this lower bound is equivalent to minimizing the KL divergence with respect to the alignment module and predictor. This yields the Routing-guided Semantic Distillation (RSD) objective:
\begin{equation}
\mathcal{L}_{\mathrm{RSD}}=
\frac{1}{BCN}
\sum_{i=1}^{B}\sum_{c=1}^{C}\sum_{n=1}^{N}
\mathrm{KL}\left(
\operatorname{sg}(\mathbf{r}_{i,c}^{n})
\parallel
\hat{\mathbf{r}}_{i,c}^{n}
\right),
\end{equation}
where $\operatorname{sg}(\cdot)$ denotes the stop-gradient operation.

\subsection{Overall Training Objective}
The overall training objective of IMRouter is :
\begin{equation}
\mathcal{L}_{\mathrm{total}} = \mathcal{L}_{\mathrm{pred}}
+
\lambda_{\mathrm{BAL}}\mathcal{L}_{\mathrm{BAL}}
+
\lambda_{\mathrm{DWGC}}\mathcal{L}_{\mathrm{DWGC}}
+
\lambda_{\mathrm{RSD}}\mathcal{L}_{\mathrm{RSD}},
\end{equation}
where $\mathcal{L}_{\mathrm{pred}}$ is the forecasting loss, and $\mathcal{L}_{\mathrm{BAL}}$ is the load-balancing regularizer. The hyperparameters $\lambda_{\mathrm{BAL}}$, $\mathcal{L}_{\mathrm{BAL}}$ and $\lambda_{\mathrm{RSD}}$ control the strength of the auxiliary objectives. Notably, routing assignments are often unstable during the early training stage, which may introduce noisy supervision into semantic distillation. We therefore activate RGSA midway through training.

\section{Experiments}

\subsection{Implementation Details}
All experiments are implemented in Python~3.10 using
PyTorch~2.1.3 and conducted on a workstation equipped
with two NVIDIA GeForce RTX~4090 GPUs. We
adopt the Adam optimizer and search the learning rate over
\(\{\text{0.0001,0.0003,0.0005,0.0008}\}\), while the batch
size is selected from \(\{\text{8,32,64,128}\}\).
The maximum number of training epochs is set to
\(\text{10,20,30,100}\), and early stopping is applied with a
patience of \(\text{5,10}\) according to the validation
MSE. To ensure reproducibility, all experiments use a fixed
random seed of \(\text{2021}\).

We implement IMRouter on top of the CALF forecasting framework \cite{DBLP:conf/aaai/LiuG0LBR0X25} by augmenting it with the proposed PATR and RGSA modules. Since CALF constructs temporal representations at the channel level, we adopt channel-wise expert routing in this implementation. To ensure a controlled comparison, we retain the same backbone, data-processing pipeline, and optimizer configuration as CALF unless otherwise stated. During training, we first optimize the model using the forecasting objective and subsequently activate the density-weighted gradient conflict regularizer after an initial warm-up stage. The expert architectures, routing mechanisms, auxiliary objectives, and their associated hyperparameters are introduced specifically for IMRouter.
\subsection{Baselines}
We compare IMRouter with representative forecasting baselines, which can be divided into two categories. The first category includes conventional deep time-series forecasting models, including PatchTST \cite{DBLP:conf/iclr/NieNSK23},  iTransformer \cite{DBLP:conf/iclr/LiuHZWWML24}, TimesNet \cite{DBLP:conf/iclr/WuHLZ0L23}, and DLinear \cite{zeng2023transformers}. The second category consists of LLM-based TSF methods, such as GPT4TS \cite{DBLP:conf/nips/ZhouNW0023}, Time-LLM \cite{DBLP:conf/iclr/0005WMCZSCLLPW24}, CALF \cite{DBLP:conf/aaai/LiuG0LBR0X25}, TimeSup \cite{DBLP:conf/www/ZhengLZXMC26} and FSCA \cite{DBLP:conf/iclr/0003LZYC25} (a state-of-the-art method for LLM-based TSF).  For a fair comparison, the backbone of LLM-based models is set to GPT-2 \cite{radford2019language}.
\subsection{Long-term Forecasting}

We evaluate IMRouter on seven widely used long-term time series forecasting benchmarks: ETTh1, ETTh2, ETTm1, ETTm2, Weather, Electricity (ECL), and Traffic \cite{DBLP:conf/iclr/WuHLZ0L23}. These datasets cover diverse application domains, including energy systems, meteorology, and transportation. Moreover, they exhibit substantial diversity in temporal resolution, dimensionality, and underlying temporal dynamics, providing a comprehensive benchmark for evaluating forecasting performance across diverse temporal scenarios. Following standard long-term forecasting settings, we fix the input sequence length at 96 and evaluate four prediction horizons, including $\{96,192,336,720\}$, to assess model performance across different forecasting lengths. We adopt Mean Squared Error (MSE) and Mean Absolute Error (MAE) as evaluation metrics. MSE emphasizes larger prediction deviations, whereas MAE provides a complementary measure of average absolute errors. Lower values indicate better forecasting performance.











\begin{table*}[t]
\centering
\setlength{\tabcolsep}{6pt}
\renewcommand{\arraystretch}{1.15}
\small

\begin{tabular}{clccccccc}
\toprule
& Models
& IMRouter
& FSCA
& Time-LLM
& GPT4TS
& PatchTST
& iTransformer
& TimesNet \\
\midrule

\multirow{3}{*}{\rotatebox{90}{Average}}
& SMAPE
& \textbf{11.77}
& \underline{11.828}
& 12.494
& 12.365
& 12.059
& 12.142
& 12.880 \\

& MASE
& \textbf{1.568}
& \underline{1.580}
& 1.731
& 1.767
& 1.623
& 1.632
& 1.836 \\

& OWA
& \textbf{0.844}
& \underline{0.850}
& 0.913
& 0.918
& 0.869
& 0.874
& 0.955 \\

\bottomrule
\end{tabular}

\caption{Average short-term forecasting performance on the M4
dataset. The results are averaged across the Yearly, Quarterly,
Monthly, and Other subsets. The best results are shown in bold,
and the second-best results are underlined.}
\label{tab:m4_results}

\end{table*}

\begin{table*}[t]
\centering

\setlength{\tabcolsep}{3pt}
\renewcommand{\arraystretch}{1.1}

\resizebox{\textwidth}{!}{%
\begin{tabular}{l|cc|cc|cc|cc|cc|cc|cc|cc|cc}
\toprule
\multirow{2}{*}{Dataset}
& \multicolumn{2}{c|}{IMRouter}
& \multicolumn{2}{c|}{FSCA}
& \multicolumn{2}{c|}{CALF}
& \multicolumn{2}{c|}{Time-LLM}
& \multicolumn{2}{c|}{GPT4TS}
& \multicolumn{2}{c|}{PatchTST}
& \multicolumn{2}{c|}{iTransformer}
& \multicolumn{2}{c|}{TimesNet}
& \multicolumn{2}{c}{DLinear} \\

& MSE & MAE
& MSE & MAE
& MSE & MAE
& MSE & MAE
& MSE & MAE
& MSE & MAE
& MSE & MAE
& MSE & MAE
& MSE & MAE \\
\midrule

ETTm1
& \textbf{0.498} & \textbf{0.457}
& 0.602 & 0.503
& \underline{0.504} & \underline{0.462}
& 0.565 & 0.491
& 0.607 & 0.498
& 0.619 & 0.496
& 0.580 & 0.497
& 0.674 & 0.535
& 0.568 & 0.500 \\

\midrule

ETTm2
& \textbf{0.292} & \textbf{0.330}
& \underline{0.302} & 0.339
& \underline{0.302} & \textbf{0.330}
& 0.303 & 0.341
& 0.303 & \underline{0.337}
& \underline{0.302} & 0.342
& 0.306 & 0.344
& 0.322 & 0.355
& 0.329 & 0.383 \\

\midrule

ETTh1
& 0.643 & 0.541
& 0.695 & 0.553
& 0.644 & 0.541
& \textbf{0.616} & \textbf{0.530}
& 0.688 & 0.554
& \underline{0.623} & \underline{0.532}
& 0.748 & 0.587
& 0.864 & 0.625
& 0.647 & 0.552 \\

\midrule

ETTh2
& \underline{0.419} & \underline{0.425}
& \textbf{0.403} & \textbf{0.416}
& \underline{0.419} & 0.427
& 0.497 & 0.465
& 0.574 & 0.495
& 0.490 & 0.459
& 0.443 & 0.443
& 0.487 & 0.468
& 0.446 & 0.461 \\

\bottomrule
\end{tabular}%
}
\label{tab:fewshot_results}

\caption{Few-shot forecasting performance on four ETT datasets.
Results are averaged over the available prediction horizons.
Lower MSE and MAE indicate better performance.
The best results are highlighted in bold, and the second-best
results are underlined.}
\end{table*}

As shown in Table.~\ref{tab:long_term_forecasting}, IMRouter achieves the lowest average MSE and MAE across seven long-term forecasting benchmarks. Compared with CALF, the strongest baseline on average, IMRouter reduces MSE and MAE by 1.4\% and 2.0\%, respectively. Relative to FSCA, Time-LLM, and GPT4TS, the corresponding reductions are 5.7\%/3.7\%, 9.9\%/7.6\%, and 4.4\%/4.1\%. IMRouter also outperforms conventional forecasting methods on average, achieving reductions of up to 12.3\% in MSE and 7.0\% in MAE. Notably, IMRouter achieves the lowest MSE on ECL and Traffic, demonstrating its effectiveness on high-dimensional forecasting benchmarks. Overall, these results highlight the effectiveness of IMRouter across diverse temporal scenarios.

\begin{table}[t]
\centering

\setlength{\tabcolsep}{4pt}
\renewcommand{\arraystretch}{1.15}
\small

\begin{tabular}{lcccc}
\toprule
\multirow{2}{*}{Variants}
& \multicolumn{2}{c}{ECL}
& \multicolumn{2}{c}{Traffic} \\
\cmidrule(lr){2-3}
\cmidrule(lr){4-5}

& MSE & MAE & MSE & MAE \\
\midrule

CALF
&0.193 &0.278 &0.459 &0.303 \\

MLP
& 0.201 & 0.288 &0.459 &0.306 \\

Vanilla MoE
& 0.186 & 0.279 & 0.443 & 0.292 \\

PATR
& 0.179 & 0.268 & 0.433 & 0.291 \\

Vanilla MoE + RGSA
& 0.180 & 0.272 & 0.430 & 0.291 \\

PATR w/o Density + RGSA
& 0.177 &0.266 &0.430 &0.289 \\

\midrule

\textbf{IMRouter}
&0.171 &0.263 &0.423 &0.285 \\

\bottomrule
\end{tabular}
\caption{Ablation results of IMRouter on ECL and Traffic.
Lower MSE and MAE indicate better performance.}
\label{tab:ablation}
\end{table}
\subsection{Short-term Forecasting}

For short-term forecasting, we evaluate IMRouter on the M4 benchmark \cite{MAKRIDAKIS2018802}, which comprises 100,000 univariate time series from diverse domains. The dataset covers six sampling frequencies: Yearly, Quarterly, Monthly, Weekly, Daily, and Hourly, with corresponding prediction horizons of 6, 8, 18, 13, 14, and 48, respectively. These subsets exhibit diverse temporal characteristics, including different seasonal patterns, sequence lengths, and sampling granularities, providing a comprehensive evaluation across diverse forecasting scenarios. Following the M4 evaluation protocol, we adopt three metrics: symmetric Mean Absolute Percentage Error (sMAPE), Mean Absolute Scaled Error (MASE), and Overall Weighted Average (OWA). Lower values indicate better performance across all three metrics.

As shown in Table.~\ref{tab:m4_results}, IMRouter achieves the best average performance across all three metrics on the M4 dataset, with sMAPE, MASE, and OWA of 11.770, 1.568, and 0.844, respectively. Compared with FSCA, the strongest competing baseline, IMRouter reduces these metrics by 0.49\%, 0.76\%, and 0.71\%, respectively. Compared with conventional forecasting models, IMRouter also achieves better performance than PatchTST, iTransformer, and TimesNet across all three metrics. These improvements suggest that IMRouter can effectively capture temporal dynamics across diverse forecasting scenarios. Overall, the results further demonstrate the effectiveness and applicability of IMRouter in short-term forecasting tasks involving diverse temporal characteristics.

\subsection{Few-shot Learning}
Few-shot learning is crucial for deploying time-series forecasting models in real-world scenarios with limited training data. To examine this capability, we evaluate IMRouter in the few-shot setting, where the model is trained using only 10\% of the available training samples. We report the average performance across four prediction horizons $\{96,192,336,720\}$. As shown in Table. 3,  IMRouter ranks first or second on six of the eight metrics, achieving the lowest MSE and MAE on ETTm1 and ETTm2. Compared with the CALF backbone, IMRouter reduces the average MSE and MAE by 0.91\% and 0.40\%, respectively. Notably, IMRouter demonstrates competitive performance across all four datasets, with particularly notable improvements on ETTm1 and ETTm2. These results suggest that IMRouter can effectively leverage limited training data to learn temporal representations, demonstrating its potential for few-shot forecasting.

\begin{figure}[t]
 \includegraphics[width=0.8\linewidth]{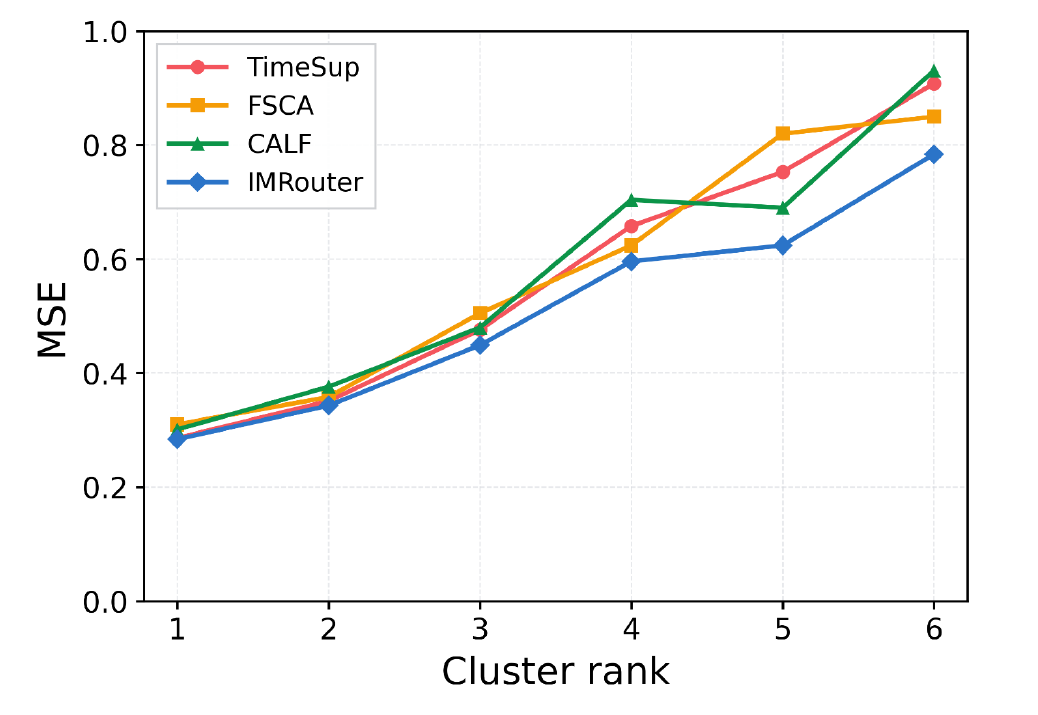}
  \caption{Pattern-wise forecasting performance of IMRouter and three baseline methods on the Traffic dataset.}
  \Description{Enjoying the baseball game from the third-base
  seats. Ichiro Suzuki preparing to bat.}
  \label{fig:patter-wise}
\end{figure}

\begin{figure}[t]
 \includegraphics[width=\linewidth]{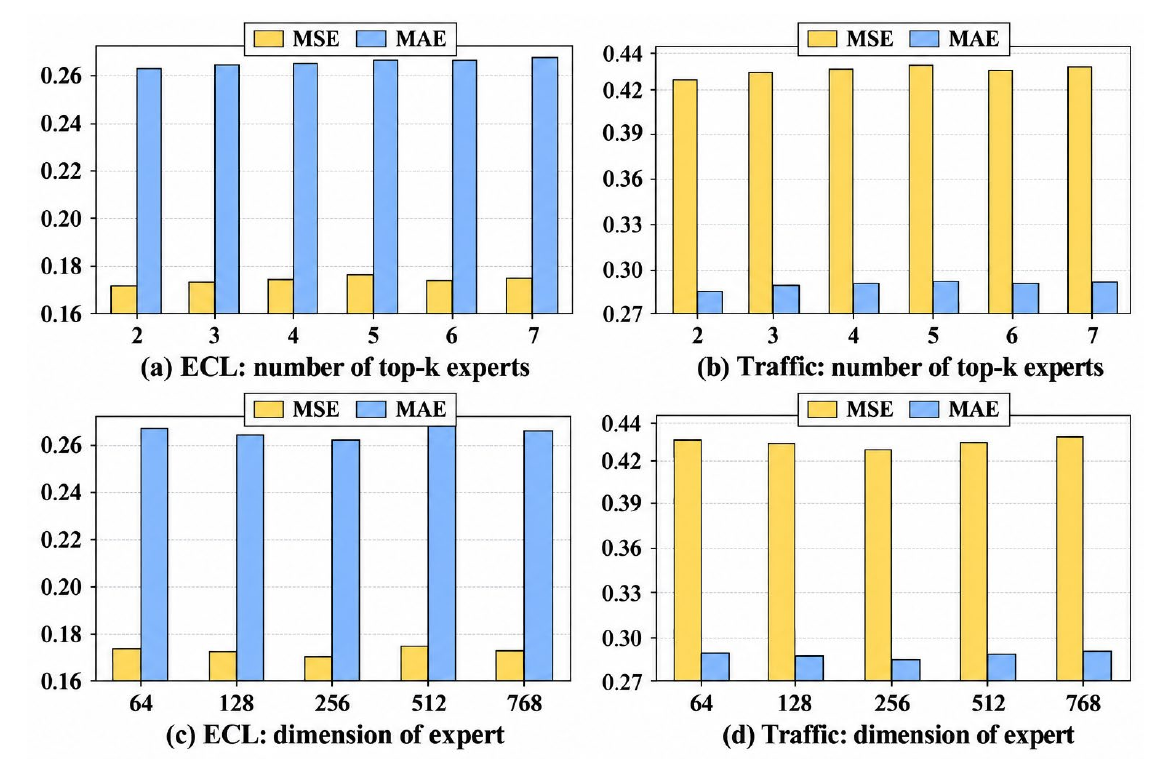}
  \caption{Hyperparameter sensitivity results on the number of top-k experts and the dimension of experts.}
  \Description{Enjoying the baseball game from the third-base
  seats. Ichiro Suzuki preparing to bat.}
  \label{sensitity}
\end{figure}
\subsection{Ablation Study}
To systematically evaluate the effectiveness of IMRouter,
we conduct ablation experiments on the ECL and Traffic
datasets. Specifically, we construct five variants: \textbf{CALF} : This variant removes PATR and RGSA. \textbf{MLP}: This variant replaces PATR with an MLP of comparable computational cost and removes RGSA. \textbf{Vanilla MoE}: This variant retains the same MoE architecture as PATR but removes gradient-guided routing regularization and RGSA. \textbf{PATR}: This variant retains the complete PATR module while removing RGSA. \textbf{Vanilla MoE + RGSA}: This variant retains RGSA but replaces PATR with a standard MoE without gradient-guided routing regularization. \textbf{PATR w/o Density + RGSA}: This variant retains gradient-conflict regularization and RGSA but removes density-aware weighting.

Table 4 presents the ablation results on ECL and Traffic. Compared with CALF and MLP, vanilla MoE achieves lower MSE on both datasets, highlighting the benefit of expert-based representations. PATR further improves performance, demonstrating the effectiveness of gradient-guided routing. Incorporating RGSA into vanilla MoE also reduces forecasting errors, indicating the benefit of routing-guided semantic alignment. Moreover, density-aware weighting further reduces MSE from 0.177 to 0.171 on ECL and from 0.430 to 0.423 on Traffic, highlighting its contribution beyond gradient-conflict regularization. Overall, IMRouter achieves the best performance on both datasets, validating the complementary contributions of PATR and RGSA.

\subsection{Pattern-wise Forecasting Performance}
To further investigate the effectiveness of IMRouter in mitigating TPI, we evaluate its forecasting performance across temporal patterns with different occurrence frequencies. Specifically, we adopt the same temporal pattern clusters and test samples as those used in Figure.~\ref{fig:pic2}(b), ensuring a consistent comparison with existing methods. The clusters are ranked by their occurrence frequencies in descending order, and the MSE of each model is calculated separately for each cluster. 

As shown in Figure.~\ref{fig:patter-wise}, forecasting errors generally increase as temporal pattern frequency decreases. Nevertheless, IMRouter achieves the lowest MSE across all six clusters, with more pronounced improvements on less frequent patterns. For instance, on Cluster 6, IMRouter achieves an MSE of approximately 0.78, compared with 0.85--0.93 for competing methods. These results demonstrate the effectiveness of IMRouter in improving forecasting accuracy on underrepresented temporal patterns, highlighting its potential to mitigate TPI.

\subsection{Hyperparameter Sensitivity}
We evaluate the sensitivity of IMRouter to the number of selected experts $k$ and the expert dimension on ECL and Traffic. Specifically, we vary one hyperparameter at a time while keeping the remaining settings fixed, and report the average MSE across four prediction horizons. The number of selected experts determines how many experts contribute to each temporal representation, while the expert dimension controls the representation capacity of individual experts. 

As illustrated in Figure.~\ref{sensitity}(a)–(b), IMRouter achieves the best performance with $k=2$ on both ECL and Traffic. Increasing the number of selected experts generally degrades performance, suggesting that sparse routing helps reduce interference between different temporal patterns. Regarding expert dimension, both datasets achieve the lowest forecasting errors at 256, while further increasing the dimension offers no additional benefits. These results highlight the importance of appropriate expert sparsity and capacity for effective temporal representation learning.

\section{Conclusion}
In this paper, we investigate Temporal Pattern Imbalance (TPI) in LLM-based time series forecasting and propose IMRouter to mitigate its adverse effects. Specifically, IMRouter employs gradient-guided expert routing to reduce optimization interference and routing-guided semantic alignment to preserve pattern-specific information. Extensive experiments demonstrate that IMRouter improves forecasting performance across diverse benchmarks, particularly for less frequent temporal patterns. These findings highlight the importance of addressing TPI for reliable LLM-based time series forecasting.

\bibliographystyle{ACM-Reference-Format}
\bibliography{sigconf}

\appendix

\section{Appendix}

\subsection{Evaluation Metrics}

For long-term and few-shot forecasting, we evaluate model
performance using Mean Squared Error (MSE) and Mean
Absolute Error (MAE). For short-term forecasting on the M4
benchmark, we follow the standard evaluation protocol and
report Symmetric Mean Absolute Percentage Error (SMAPE),
Mean Absolute Scaled Error (MASE), and Overall Weighted
Average (OWA). Lower values indicate better forecasting
performance for all metrics.

Let \(B\), \(C\), and \(H\) denote the number of forecasting
windows, the number of variables, and the prediction horizon,
respectively. Let \(y_{i,c,h}\) and \(\hat{y}_{i,c,h}\) denote
the ground-truth and predicted values at the \(h\)-th
forecasting step for the \(c\)-th variable of the \(i\)-th
window. MSE and MAE are defined as
\begin{align}
\mathrm{MSE}
&=
\frac{1}{BCH}
\sum_{i=1}^{B}
\sum_{c=1}^{C}
\sum_{h=1}^{H}
\left(
y_{i,c,h}-\hat{y}_{i,c,h}
\right)^2,
\label{eq:metric_mse}\\
\mathrm{MAE}
&=
\frac{1}{BCH}
\sum_{i=1}^{B}
\sum_{c=1}^{C}
\sum_{h=1}^{H}
\left|
y_{i,c,h}-\hat{y}_{i,c,h}
\right|.
\label{eq:metric_mae}
\end{align}

For the M4 dataset, let \(N\) denote the number of time
series in a subset. For the \(i\)-th series, \(H_i\) denotes
its forecasting horizon, \(T_i\) denotes the length of its
training sequence, and \(s_i\) denotes its seasonal
periodicity. The series-level SMAPE is computed as
\begin{equation}
\mathrm{SMAPE}_i
=
\frac{200}{H_i}
\sum_{h=1}^{H_i}
\frac{
\left|
y_{i,h}-\hat{y}_{i,h}
\right|
}{
\left|y_{i,h}\right|
+
\left|\hat{y}_{i,h}\right|
}.
\label{eq:metric_smape}
\end{equation}

MASE scales the forecasting error by the in-sample error of
a seasonal naive forecast:
\begin{equation}
\mathrm{MASE}_i
=
\frac{
\displaystyle
\frac{1}{H_i}
\sum_{h=1}^{H_i}
\left|
y_{i,h}-\hat{y}_{i,h}
\right|
}{
\displaystyle
\frac{1}{T_i-s_i}
\sum_{t=s_i+1}^{T_i}
\left|
y_{i,t}-y_{i,t-s_i}
\right|
}.
\label{eq:metric_mase}
\end{equation}

The subset-level scores are obtained by averaging the
series-level metrics:
\begin{equation}
\mathrm{SMAPE}
=
\frac{1}{N}
\sum_{i=1}^{N}\mathrm{SMAPE}_i,
\qquad
\mathrm{MASE}
=
\frac{1}{N}
\sum_{i=1}^{N}\mathrm{MASE}_i.
\label{eq:metric_average}
\end{equation}

Finally, OWA measures the relative performance against the
Naive2 forecasting method:
\begin{equation}
\mathrm{OWA}
=
\frac{1}{2}
\left(
\frac{\mathrm{SMAPE}}
     {\mathrm{SMAPE}_{\mathrm{Naive2}}}
+
\frac{\mathrm{MASE}}
     {\mathrm{MASE}_{\mathrm{Naive2}}}
\right).
\label{eq:metric_owa}
\end{equation}

Here, \(\mathrm{SMAPE}_{\mathrm{Naive2}}\) and
\(\mathrm{MASE}_{\mathrm{Naive2}}\) denote the corresponding
scores obtained by the Naive2 benchmark. An OWA value below
\(1\) indicates that the evaluated method outperforms Naive2
on average.

\subsection{Full Results}
\subsubsection{Long-term Forecasting}
Table.~\ref{tab:main_results_updated} reports the complete long-term forecasting results across all datasets and prediction horizons $H\in\{96,192,336,720\}$. In addition to the averaged results presented in the main paper, we provide horizon-specific MSE and MAE values to enable a more detailed comparison of model performance and facilitate result verification and reproducibility.

\subsubsection{Short-term Forecasting}
To provide a fine-grained evaluation beyond the aggregated results in the main paper, Table.~\ref{tab:m4_detailed_results} reports the complete short-term forecasting performance on the M4 benchmark. We present sMAPE, MASE, and OWA for each frequency-specific subset, enabling a detailed assessment of model robustness across time series with different sampling frequencies, seasonal characteristics, and forecasting horizons.

\subsubsection{Few-shot Forecasting}
Tab.~\ref{tab:full_ett_results} presents the complete few-shot forecasting results using only 10\% of the available training data. In addition to the averaged performance reported in the main paper, we provide horizon-specific MSE and MAE on the four ETT benchmarks to examine model behavior under different forecasting lengths. These detailed results enable a more comprehensive assessment of data efficiency and robustness in limited-data scenarios.
\begin{table*}[t]
\centering
\small
\setlength{\tabcolsep}{3.2pt}
\renewcommand{\arraystretch}{1.1}
\begin{tabular}{l|c|cc|cc|cc|cc|cc|cc|cc|cc|cc}
\toprule
\multirow{2}{*}{Dataset} & \multirow{2}{*}{Pred. Len.}
& \multicolumn{2}{c|}{IMRouter}
& \multicolumn{2}{c|}{TimeSup$^\dagger$}
& \multicolumn{2}{c|}{FSCA$^\dagger$}
& \multicolumn{2}{c|}{CALF}
& \multicolumn{2}{c|}{Time-LLM}
& \multicolumn{2}{c|}{GPT4TS}
& \multicolumn{2}{c|}{PatchTST}
& \multicolumn{2}{c|}{iTransformer}
& \multicolumn{2}{c}{TimesNet} \\
\cline{3-20}
& & MSE & MAE & MSE & MAE & MSE & MAE & MSE & MAE & MSE & MAE & MSE & MAE & MSE & MAE & MSE & MAE & MSE & MAE \\
\midrule

\multirow{5}{*}{ETTm1}
& 96  & 0.317 & 0.348 & 0.338 & 0.370 & 0.640 & 0.496 & 0.324 & 0.353 & 0.331 & 0.370 & 0.330 & 0.365 & 0.324 & 0.365 & 0.341 & 0.376 & 0.334 & 0.375 \\
& 192 & 0.369 & 0.375 & 0.385 & 0.394 & 0.374 & 0.391 & 0.369 & 0.376 & 0.368 & 0.383 & 0.368 & 0.383 & 0.372 & 0.392 & 0.381 & 0.395 & 0.407 & 0.413 \\
& 336 & 0.399 & 0.396 & 0.411 & 0.414 & 0.409 & 0.412 & 0.407 & 0.399 & 0.411 & 0.413 & 0.401 & 0.404 & 0.398 & 0.409 & 0.419 & 0.419 & 0.415 & 0.422 \\
& 720 & 0.473 & 0.437 & 0.476 & 0.449 & 0.471 & 0.449 & 0.472 & 0.436 & 0.461 & 0.439 & 0.461 & 0.439 & 0.457 & 0.444 & 0.486 & 0.456 & 0.486 & 0.461 \\
& Avg & 0.389 & 0.389 & 0.402 & 0.407 & 0.473 & 0.437 & 0.393 & 0.391 & 0.392 & 0.403 & 0.390 & 0.401 & 0.388 & 0.402 & 0.407 & 0.411 & 0.410 & 0.418 \\
\midrule

\multirow{5}{*}{ETTm2}
& 96  & 0.189 & 0.267 & 0.180 & 0.263 & 0.181 & 0.263 & 0.178 & 0.255 & 0.181 & 0.267 & 0.178 & 0.264 & 0.186 & 0.268 & 0.184 & 0.267 & 0.190 & 0.267 \\
& 192 & 0.256 & 0.311 & 0.244 & 0.304 & 0.246 & 0.303 & 0.243 & 0.298 & 0.225 & 0.311 & 0.245 & 0.306 & 0.246 & 0.305 & 0.253 & 0.312 & 0.253 & 0.307 \\
& 336 & 0.320 & 0.348 & 0.308 & 0.346 & 0.303 & 0.340 & 0.303 & 0.338 & 0.315 & 0.352 & 0.309 & 0.346 & 0.311 & 0.348 & 0.315 & 0.352 & 0.321 & 0.349 \\
& 720 & 0.403 & 0.396 & 0.405 & 0.412 & 0.399 & 0.398 & 0.403 & 0.396 & 0.404 & 0.401 & 0.404 & 0.401 & 0.418 & 0.413 & 0.412 & 0.406 & 0.418 & 0.405 \\
& Avg & 0.292 & 0.330 & 0.284 & 0.329 & 0.282 & 0.326 & 0.282 & 0.322 & 0.291 & 0.333 & 0.284 & 0.329 & 0.290 & 0.333 & 0.291 & 0.334 & 0.295 & 0.332 \\
\midrule

\multirow{5}{*}{ETTh1}
& 96  & 0.378 & 0.395 & 0.386 & 0.404 & 0.370 & 0.397 & 0.386 & 0.410 & 0.394 & 0.407 & 0.377 & 0.398 & 0.377 & 0.397 & 0.395 & 0.410 & 0.389 & 0.412 \\
& 192 & 0.434 & 0.429 & 0.447 & 0.439 & 0.420 & 0.425 & 0.431 & 0.430 & 0.446 & 0.437 & 0.439 & 0.427 & 0.426 & 0.432 & 0.449 & 0.441 & 0.439 & 0.442 \\
& 336 & 0.477 & 0.451 & 0.429 & 0.436 & 0.494 & 0.461 & 0.466 & 0.470 & 0.485 & 0.459 & 0.480 & 0.447 & 0.469 & 0.457 & 0.492 & 0.465 & 0.494 & 0.471 \\
& 720 & 0.480 & 0.472 & 0.529 & 0.497 & 0.466 & 0.470 & 0.473 & 0.466 & 0.493 & 0.484 & 0.496 & 0.477 & 0.519 & 0.504 & 0.522 & 0.504 & 0.518 & 0.494 \\
& Avg & 0.442 & 0.437 & 0.460 & 0.450 & 0.429 & 0.436 & 0.441 & 0.439 & 0.454 & 0.447 & 0.448 & 0.437 & 0.448 & 0.447 & 0.464 & 0.455 & 0.460 & 0.455 \\
\midrule

\multirow{5}{*}{ETTh2}
& 96  & 0.289 & 0.338 & 0.309 & 0.354 & 0.389 & 0.343 & 0.297 & 0.351 & 0.298 & 0.346 & 0.294 & 0.347 & 0.309 & 0.359 & 0.300 & 0.350 & 0.337 & 0.371 \\
& 192 & 0.373 & 0.388 & 0.392 & 0.406 & 0.374 & 0.397 & 0.369 & 0.388 & 0.389 & 0.401 & 0.386 & 0.404 & 0.381 & 0.407 & 0.382 & 0.400 & 0.405 & 0.415 \\
& 336 & 0.415 & 0.424 & 0.434 & 0.441 & 0.383 & 0.409 & 0.422 & 0.426 & 0.431 & 0.438 & 0.423 & 0.435 & 0.412 & 0.429 & 0.424 & 0.432 & 0.450 & 0.449 \\
& 720 & 0.417 & 0.436 & 0.472 & 0.473 & 0.443 & 0.409 & 0.421 & 0.450 & 0.431 & 0.446 & 0.424 & 0.447 & 0.436 & 0.456 & 0.426 & 0.445 & 0.435 & 0.448 \\
& Avg & 0.372 & 0.397 & 0.402 & 0.419 & 0.372 & 0.390 & 0.377 & 0.404 & 0.387 & 0.408 & 0.382 & 0.408 & 0.384 & 0.413 & 0.383 & 0.407 & 0.407 & 0.421 \\
\midrule

\multirow{5}{*}{Weather}
& 96  & 0.163 & 0.199 & 0.176 & 0.215 & 0.178 & 0.217 & 0.163 & 0.203 & 0.195 & 0.234 & 0.182 & 0.224 & 0.176 & 0.218 & 0.175 & 0.216 & 0.169 & 0.219 \\
& 192 & 0.214 & 0.246 & 0.221 & 0.256 & 0.227 & 0.259 & 0.210 & 0.248 & 0.239 & 0.269 & 0.232 & 0.263 & 0.221 & 0.256 & 0.225 & 0.258 & 0.225 & 0.365 \\
& 336 & 0.271& 0.288 & 0.281 & 0.297 & 0.280 & 0.297 & 0.266 & 0.290 & 0.291 & 0.306 & 0.283 & 0.299 & 0.280 & 0.298 & 0.280 & 0.298 & 0.282 & 0.304 \\
& 720 & 0.353 & 0.342 & 0.357 & 0.348 & 0.356 & 0.348 & 0.347 & 0.344 & 0.366 & 0.353 & 0.360 & 0.349 & 0.356 & 0.348 & 0.361 & 0.351 & 0.359 & 0.354 \\
& Avg & 0.250 & 0.269 & 0.259 & 0.279 & 0.260 & 0.280 & 0.251 & 0.271 & 0.273 & 0.290 & 0.264 & 0.284 & 0.258 & 0.280 & 0.260 & 0.281 & 0.259 & 0.285 \\
\midrule

\multirow{5}{*}{Traffic}
& 96  & 0.394& 0.270 & 0.419 & 0.274 & 0.433 & 0.281 & 0.433 & 0.292 & 0.487 & 0.307 & 0.468 & 0.308 & 0.459 & 0.298 & 0.392 & 0.268 & 0.589 & 0.315 \\
& 192 & 0.414 & 0.280 & 0.431 & 0.283 & 0.446 & 0.287 & 0.451 & 0.297 & 0.499 & 0.318 & 0.477 & 0.311 & 0.469 & 0.301 & 0.412 & 0.277 & 0.617 & 0.324 \\
& 336 & 0.427 & 0.285 & 0.444 & 0.289 & 0.460 & 0.293 & 0.467 & 0.302 & 0.516 & 0.329 & 0.489 & 0.317 & 0.483 & 0.307 & 0.424 & 0.283 & 0.635 & 0.338 \\
& 720 & 0.457 & 0.304 & 0.474 & 0.307 & 0.491 & 0.311 & 0.485 & 0.323 & 0.535 & 0.343 & 0.522 & 0.333 & 0.517 & 0.326 & 0.460 & 0.301 & 0.660 & 0.349 \\
& Avg & 0.423 & 0.285 & 0.440 & 0.288 & 0.457 & 0.293 & 0.459 & 0.303 & 0.509 & 0.324 & 0.489 & 0.317 & 0.482 & 0.308 & 0.422 & 0.282 & 0.625 & 0.331 \\
\midrule

\multirow{5}{*}{ECL}
& 96  & 0.141 & 0.235 & 0.166 & 0.255 & 0.170 & 0.255 & 0.167 & 0.254 & 0.169 & 0.259 & 0.186 & 0.272 & 0.180 & 0.273 & 0.148 & 0.240 & 0.169 & 0.272 \\
& 192 & 0.158 & 0.250 & 0.176 & 0.247 & 0.178 & 0.266 & 0.178 & 0.264 & 0.173 & 0.266 & 0.189 & 0.277 & 0.187 & 0.280 & 0.165 & 0.256 & 0.187 & 0.289 \\
& 336 & 0.176& 0.269 & 0.176 & 0.266 & 0.195 & 0.280 & 0.196 & 0.282 & 0.197 & 0.289 & 0.205 & 0.291 & 0.204 & 0.296 & 0.178 & 0.270 & 0.199 & 0.299 \\
& 720 & 0.211 & 0.298 & 0.235 & 0.319 & 0.233 & 0.312 & 0.233 & 0.312 & 0.249 & 0.328 & 0.244 & 0.324 & 0.246 & 0.328 & 0.209 & 0.299 & 0.232 & 0.329 \\
& Avg & 0.177 & 0.263 & 0.192 & 0.281 & 0.194 & 0.278 & 0.193 & 0.278 & 0.197 & 0.285 & 0.206 & 0.291 & 0.204 & 0.294 & 0.175 & 0.266 & 0.197 & 0.297 \\
\bottomrule
\end{tabular}
\caption{Multivariate long-term forecasting results.}
\label{tab:main_results_updated}
\end{table*}

\begin{table*}[t]
\centering

\label{tab:m4_detailed_results}

\setlength{\tabcolsep}{5pt}
\renewcommand{\arraystretch}{1.1}
\small

\resizebox{\textwidth}{!}{%
\begin{tabular}{c|l|cccccccc}
\toprule
& Models
& IMRouter
& FSCA
& CALF
& Time-LLM
& GPT4TS
& PatchTST
& iTransformer
& TimesNet \\
\midrule

\multirow{3}{*}{\rotatebox{90}{Yearly}}
& sMAPE
& 13.406 & 13.288 & 13.351 & 13.750
& 14.822 & 13.477 & 13.652 & 15.378 \\

& MASE
& 3.001 & 2.974 & 3.003 & 3.055
& 3.618 & 3.019 & 3.095 & 3.554 \\

& OWA
& 0.788 & 0.781 & 0.786 & 0.805
& 0.909 & 0.792 & 0.807 & 0.918 \\

\midrule

\multirow{3}{*}{\rotatebox{90}{Quarterly}}
& sMAPE
& 9.986 & 10.037 & 9.990 & 10.671
& 10.411 & 10.380 & 10.353 & 10.465 \\

& MASE
& 1.161 & 1.174 & 1.164 & 1.276
& 1.232 & 1.233 & 1.209 & 1.227 \\

& OWA
& 0.877 & 0.884 & 0.878 & 0.950
& 0.922 & 0.921 & 0.911 & 0.923 \\

\midrule

\multirow{3}{*}{\rotatebox{90}{Monthly}}
& sMAPE
& 12.623 & 12.762 & 12.643 & 13.416
& 12.902 & 12.959 & 13.079 & 13.513 \\

& MASE
& 0.920 & 0.947 & 0.922 & 1.045
& 0.956 & 0.970 & 0.974 & 1.039 \\

& OWA
& 0.870 & 0.897 & 0.872 & 0.957
& 0.897 & 0.905 & 0.911 & 0.957 \\

\midrule

\multirow{3}{*}{\rotatebox{90}{Others}}
& sMAPE
& 4.63 & 4.761 & 4.552 & 4.973
& 5.294 & 4.952 & 4.780 & 6.913 \\

& MASE
& 3.164 & 3.207 & 3.092 & 3.412
& 3.610 & 3.347 & 3.231 & 4.507 \\

& OWA
& 0.986 & 1.007 & 0.967 & 1.053
& 1.126 & 1.049 & 1.012 & 1.438 \\

\bottomrule
\end{tabular}%
}
\caption{Short-term forecasting performance on the M4 dataset
across four subsets.}
\label{tab:m4_detailed_results}
\end{table*}

\begin{table*}[t]
\centering

\label{tab:fewshot_updated}
\setlength{\tabcolsep}{2.5pt}
\renewcommand{\arraystretch}{1.05}
\resizebox{\textwidth}{!}{%
\begin{tabular}{l|c|cc|cc|cc|cc|cc|cc|cc|cc|cc}
\toprule
\multirow{2}{*}{Dataset}
& \multirow{2}{*}{Pred. Len.}
& \multicolumn{2}{c|}{IMRouter}
& \multicolumn{2}{c|}{FSCA}
& \multicolumn{2}{c|}{CALF}
& \multicolumn{2}{c|}{Time-LLM}
& \multicolumn{2}{c|}{GPT4TS}
& \multicolumn{2}{c|}{PatchTST}
& \multicolumn{2}{c|}{iTransformer}
& \multicolumn{2}{c|}{TimesNet}
& \multicolumn{2}{c}{DLinear} \\
& & MSE & MAE & MSE & MAE & MSE & MAE
& MSE & MAE & MSE & MAE & MSE & MAE
& MSE & MAE & MSE & MAE & MSE & MAE \\
\midrule

\multirow{5}{*}{ETTm1}
& 96
& 0.452 & 0.432 & 0.573 & 0.494
& 0.468 & 0.445 & 0.571 & 0.494
& 0.613 & 0.494 & 0.654 & 0.509
& 0.571 & 0.487 & 0.585 & 0.504
& 0.552 & 0.488 \\
& 192
& 0.454 & 0.431 & 0.574 & 0.487
& 0.479 & 0.446 & 0.509 & 0.457
& 0.596 & 0.489 & 0.617 & 0.481
& 0.557 & 0.480 & 0.609 & 0.516
& 0.546 & 0.487 \\
& 336
& 0.510 & 0.465 & 0.618 & 0.509
& 0.499 & 0.463 & 0.576 & 0.498
& 0.595 & 0.499 & 0.575 & 0.488
& 0.582 & 0.503 & 0.736 & 0.573
& 0.567 & 0.501 \\
& 720
& 0.578 & 0.502 & 0.644 & 0.521
& 0.572 & 0.496 & 0.606 & 0.517
& 0.623 & 0.511 & 0.631 & 0.507
& 0.612 & 0.518 & 0.768 & 0.548
& 0.606 & 0.523 \\
& Avg
& 0.591 & 0.502 & 0.602 & 0.503
& 0.504 & 0.462 & 0.565 & 0.491
& 0.607 & 0.498 & 0.619 & 0.496
& 0.580 & 0.497 & 0.674 & 0.535
& 0.568 & 0.500 \\
\midrule

\multirow{5}{*}{ETTm2}
& 96
& 0.195 & 0.278 & 0.189 & 0.268
& 0.190 & 0.268 & 0.190 & 0.271
& 0.186 & 0.267 & 0.193 & 0.274
& 0.195 & 0.277 & 0.215 & 0.289
& 0.225 & 0.320 \\
& 192
& 0.268 & 0.324 & 0.266 & 0.320
& 0.257 & 0.311 & 0.258 & 0.314
& 0.253 & 0.308 & 0.254 & 0.313
& 0.262 & 0.318 & 0.271 & 0.325
& 0.291 & 0.362 \\
& 336
& 0.328 & 0.358 & 0.317 & 0.348
& 0.323 & 0.334 & 0.323 & 0.357
& 0.333 & 0.354 & 0.326 & 0.361
& 0.325 & 0.355 & 0.330 & 0.357
& 0.354 & 0.403 \\
& 720
& 0.451 & 0.429 & 0.437 & 0.420
& 0.441 & 0.410 & 0.442 & 0.421
& 0.439 & 0.419 & 0.436 & 0.419
& 0.444 & 0.425 & 0.473 & 0.448
& 0.447 & 0.448 \\
& Avg
& 0.310 & 0.347 & 0.302 & 0.339
& 0.302 & 0.330 & 0.303 & 0.341
& 0.303 & 0.337 & 0.302 & 0.342
& 0.306 & 0.344 & 0.322 & 0.355
& 0.329 & 0.383 \\
\midrule

\multirow{5}{*}{ETTh1}
& 96
& 0.460 & 0.452 & 0.523 & 0.473
& 0.468 & 0.457 & 0.543 & 0.486
& 0.462 & 0.446 & 0.455 & 0.444
& 0.623 & 0.530 & 0.856 & 0.626
& 0.589 & 0.515 \\
& 192
& 0.550 & 0.502 & 0.552 & 0.491
& 0.550 & 0.501 & 0.566 & 0.499
& 0.549 & 0.495 & 0.531 & 0.488
& 0.694 & 0.562 & 0.786 & 0.587
& 0.631 & 0.540 \\
& 336
& 0.581 & 0.522 & 0.615 & 0.525
& 0.581 & 0.521 & 0.606 & 0.521
& 0.628 & 0.539 & 0.624 & 0.539
& 0.768 & 0.593 & 0.943 & 0.649
& 0.659 & 0.555 \\

& 720
& 0.978 & 0.687
& 1.089 & 0.724
& 0.978 & 0.685
& 0.750 & 0.615
& 1.113 & 0.738
& 0.881 & 0.657
& 0.909 & 0.662
& 0.872 & 0.639
& 0.710 & 0.600 \\
& Avg
& 0.643 & 0.541 & 0.563 & 0.496
& 0.644 & 0.541 & 0.572 & 0.502
& 0.546 & 0.493 & 0.537 & 0.490
& 0.695 & 0.562 & 0.862 & 0.621
& 0.626 & 0.537 \\
\midrule

\multirow{5}{*}{ETTh2}
& 96
& 0.317 & 0.358 & 0.305 & 0.347
& 0.314 & 0.360 & 0.319 & 0.358
& 0.321 & 0.357 & 0.315 & 0.361
& 0.340 & 0.377 & 0.367 & 0.401
& 0.361 & 0.407 \\
& 192
& 0.405 & 0.411 & 0.392 & 0.399
& 0.404 & 0.411 & 0.436 & 0.435
& 0.399 & 0.403 & 0.467 & 0.441
& 0.438 & 0.432 & 0.508 & 0.475
& 0.445 & 0.453 \\
& 336
& 0.454 & 0.451 & 0.460 & 0.456
& 0.458 & 0.452 & 0.516 & 0.481
& 0.562 & 0.499 & 0.553 & 0.490
& 0.495 & 0.474 & 0.558 & 0.504
& 0.521 & 0.508 \\

& 720
& 0.500 & 0.487
& 0.454 & 0.462
& 0.502 & 0.487
& 0.716 & 0.587
& 1.013 & 0.721
& 0.624 & 0.545
& 0.501 & 0.488
& 0.514 & 0.494
& 0.457 & 0.475 \\
& Avg
& 0.419 & 0.425 & 0.386 & 0.401
& 0.419 & 0.427 & 0.424 & 0.425
& 0.427 & 0.420 & 0.445 & 0.431
& 0.424 & 0.428 & 0.478 & 0.460
& 0.442 & 0.456 \\
\bottomrule
\end{tabular}%
}

\caption{Few-shot forecasting results on four ETT benchmarks.
}
\end{table*}


\end{document}